\documentclass[lettersize,journal]{IEEEtran}
\usepackage{amsmath,amsfonts}
\usepackage{algorithmic}
\usepackage{array}
\usepackage[caption=false,font=normalsize,labelfont=sf,textfont=sf]{subfig}
\usepackage{textcomp}
\usepackage{stfloats}
\usepackage{url}
\usepackage{verbatim}
\usepackage{graphicx}
\usepackage{color} 
\usepackage{multirow}
\usepackage{colortbl}
\usepackage{booktabs} 
\usepackage{algorithm}
\usepackage{algorithmic}
\usepackage{multirow}
\usepackage[table]{xcolor}
\definecolor{mygray}{gray}{0.8}
\definecolor{Gray}{gray}{0.1}
\definecolor{baselinecolor}{gray}{.9}
\usepackage{graphicx}

\usepackage{newtxtext}
\usepackage{newtxmath}

\def\BibTeX{{\rm B\kern-.05em{\sc i\kern-.025em b}\kern-.08em
    T\kern-.1667em\lower.7ex\hbox{E}\kern-.125emX}}
\usepackage{balance}

\usepackage{biblatex}
\begin{document}

\title{Mastering Negation: Boosting Grounding Models via Grouped Opposition-Based Learning}

\author{
Zesheng~Yang,
Xi~Jiang,
Bingzhang~Hu,
Weili~Guan,
Runmin~Cong,
Guo-Jun~Qi,~\IEEEmembership{Fellow,~IEEE},
and~Feng~Zheng,~\IEEEmembership{Member,~IEEE}
\thanks{Zesheng Yang, Xi Jiang, and Feng Zheng are with the Department of Computer Science, Southern University of Science and Technology, Shenzhen, China (e-mail: 12231140@mail.sustech.edu.cn; jiangx2020@mail.sustech.edu.cn; f.zheng@ieee.org).}

\thanks{Bingzhang Hu is with the Vision Intelligence Center, Hefei CAS Dihuge Automation Co., LTD, Hefei, China (e-mail: hubingzhang@dihuge.com).}

\thanks{Weili Guan is with the School of Information Science and Technology, Harbin Institute of Technology (Shenzhen), China (e-mail: guanweili@hit.edu.cn).}

\thanks{Runmin Cong is with the School of Control Science and Engineering, Shandong University, Jinan, China, and also with the Key Laboratory of Machine Intelligence and System Control, Ministry of Education, China (e-mail: rmcong@sdu.edu.cn).}

\thanks{Guo-Jun Qi is with the Research Center for Industries of the Future and the School of Engineering, Westlake University, Hangzhou, China, and also with OPPO Research, Seattle, WA, USA (e-mail: guojunq@gmail.com).}

\thanks{Corresponding authors: Feng Zheng (zhengf@sustech.edu.cn) and Bingzhang Hu (hubingzhang@dihuge.com).}
}

 \markboth{IEEE Transactions on Neural Networks and Learning Systems}
{Shell \MakeLowercase{\textit{et al.}}: A Sample Article Using IEEEtran.cls for IEEE Journals}

\maketitle

\begin{abstract}
Current vision-language detection and grounding models predominantly focus on processing prompts with positive-semantics and often struggle to understand and ground complex prompts accurately, particularly when these prompts contain negative-semantics.
A significant reason for this limitation is the lack of training on high-quality, discriminative negative samples and negative-semantics samples. 
To efficiently enhance the precise visual localization capabilities of vision-language models, we first introduce \textit{D-Negation}, a new dataset containing objects with both positive and negative semantic descriptions. 
Due to the widespread nature of negative logic, we propose a novel grouped opposition-based learning mechanism that extracts comprehensive negation reasoning patterns from limited samples. 
This learning strategy organizes opposing semantics from \textit{D-Negation} to establish two loss functions that enable the model to understand negation and qualifiers.
We apply our dataset and approach to the state-of-the-art language-based grounding model. 
With tuning less than 10\% of the parameters, we observe a maximum increase of 4.4 and 5.7 mAP in positive and negative semantic evaluations, respectively.

\end{abstract}


\section{Introduction}

\label{sec:intro}

Visual Grounding (VG) \cite{Qiao_Deng_Wu_2021, Deng_Yang_Chen_Zhou_Li_2021,yu2016modeling,tnnv,hivg} is a fundamental task within the Vision-Language domain aimed at accurately identifying and locating objects in an image based on natural language descriptions. 
This capability is essential for a wide range of applications, such as robot navigation \cite{fu2022coupling}, where precise object localization is crucial for interaction with the environment, and visual question answering \cite{Sun_Guo_Zhang_Li_2022}, which relies on understanding and linking visual content with textual queries. 
A robust model must be capable of understanding various forms of expressions, integrating this understanding with the image content, and ultimately providing the precise location of the described object within the image.

\begin{figure}[tbh]
    \centering
    \includegraphics[width=\linewidth]{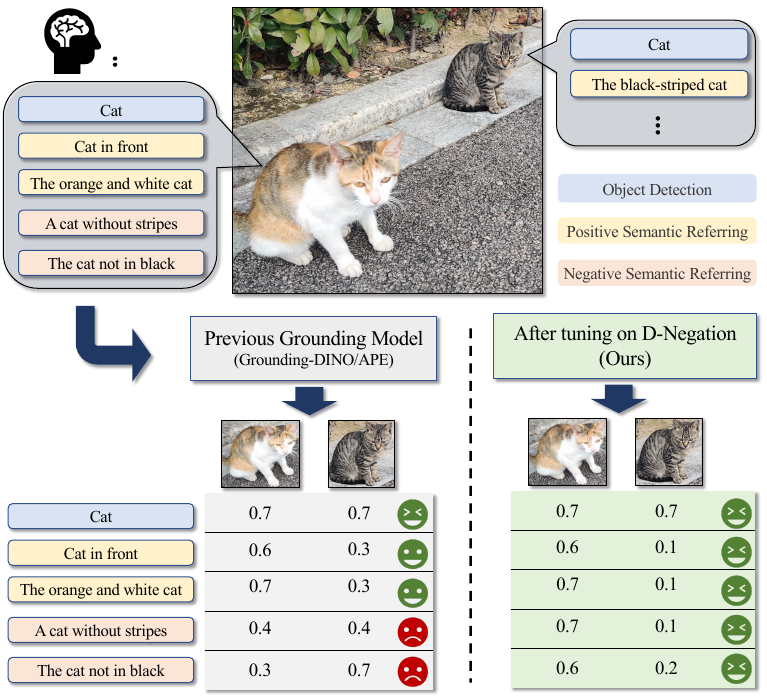}
    \caption{Intuition of positive and negative semantic referring. Besides object names and positive logic descriptions, humans can refer to objects with negative-semantic. For instance, when unsure how to describe the color of the calico cat, one might say, ``the cat not in black.'' Existing grounding models struggle to understand such referring and may even provide opposite localization results. By further fine-tuning, our method can efficiently enhance the grounding ability to complex referring. 
    }
    \label{fig:intuition}
    \vspace{-0.8em}
\end{figure}

In addition to directly specifying the category of an object, humans often use detailed descriptions when referring to objects~\cite{zeijlstra2007negation}. 
This leads to two key tasks commonly addressed in VG: Open-Vocabulary Object Detection (OVD) and Referring Expression Comprehension (REC). 
As illustrated in Fig.\ref{fig:intuition}, humans can describe the same object using not only positive but also negative-semantics when humans need to refer to a specific object in an image. 
Conventional methodologies \cite{p1,p2,p3,p4} for VG typically employ Vision-Language alignment techniques to detect and ground objects. 
While these methods have demonstrated success in standard scenarios, they frequently overlook more complex aspects of natural language, particularly negative-semantics.
We have observed that existing VG models ~\cite{p1,p2,p3} may ignore negation words, resulting in completely erroneous outputs.

Negative-semantic is a fundamental aspect of natural language that enables more nuanced and precise communication \cite{lj2,lj3}. 
It allows for expressing contrasts, exclusions, and conditions that are not straightforwardly positive \cite{lj1}.

Essentially, grounding negative-semantics descriptions poses two key challenges:
\begin{enumerate}
\item \textbf{Qualifier Comprehension:} Prompts often involve complex qualifiers such as color, spatial position, quantity, or physical state. Accurately resolving these fine-grained attributes requires the model to capture subtle semantic distinctions and align them with precise visual cues. However, current models frequently conflate similar attributes or overlook contextual modifiers, leading to ambiguous or incorrect grounding.
\item \textbf{Negation Comprehension:} Prompts may explicitly include negative semantics, expressed through terms such as \textit{not}, \textit{no}, or \textit{without}. Understanding negation is inherently difficult because it requires reasoning over the absence or exclusion of visual evidence, rather than its presence. Most existing methods, trained predominantly on positive descriptions, fail to generalize under such settings and often misinterpret negated instructions, resulting in false grounding.
\end{enumerate}

The model needs first to understand the meaning of the modifiers and then their negation. Enhancing the ability of existing REC models to handle negative semantics efficiently is therefore a significant research problem. Moreover, as previously discussed, strengthening negation understanding also improves comprehension of modifiers, thereby benefiting positive semantics as well.  
To address these challenges, we introduce both a novel dataset and an efficient fine-tuning mechanism tailored to negative semantics, as detailed below.

\paragraph{\textit{D-Negation} Dataset.} 
We propose the first visual grounding dataset that contains both positive and negative semantic descriptions with multiple attributes.
Existing visual grounding (VG) datasets \cite{da1, da2, da4}, textual labels typically correspond to individual words or simple phrases within images. These labels often provide only the names of objects, as seen in LVIS~\cite{gupta2019lvis} and Object365~\cite{shao2019objects365}, or describe objects exclusively with affirmative descriptions, as seen in Flickr30K~\cite{da3} and GQA~\cite{hudson2019gqa}. 
We contend that enhancing grounding models' understanding of negation requires a specialized dataset, so we developed a pipeline that utilizes the advanced language capabilities of the contemporary multimodal large language model (MLLM) \cite{tnn3,tnn2,tnn1}  to generate both negative and positive referring expressions for various attributes based on existing object detection annotations. 
We consider expressing negation to be a straightforward task for MLLM  (e.g., GPT-4V)\cite{tnn3}, and this has proven to be true. After filtering, we have developed a dataset with opposite referring expressions incorporating negative semantics, which serves as the foundation for our proposed fine-tuning strategy.
\paragraph{\textit{GOBL} Fine-tuning Mechanism.}  
Humans often interpret negative semantics by implicitly contrasting them with their positive counterparts. For instance, when encountering ``a cat without stripes,'' one naturally thinks of ``a cat with stripes'' before locating other cats in the image. Inspired by this reasoning process, we propose a fine-tuning mechanism called \textit{Grouped Opposition-Based Learning (GOBL)}, which enhances a model’s ability to comprehend and ground negative semantics through opposition-based learning.  
Considering that existing grounding models are trained on large-scale datasets ($\sim$20M), while \textit{D-Negation} is relatively small ($\sim$14k), we design an efficient fine-tuning strategy that focuses on the language–vision fusion module, where models often struggle to associate negative expressions with image objects.
Our training mechanism’s effectiveness was validated on two state-of-the-art grounding models. With less than 10\% of parameters tuned, our approach achieves up to a 5.7 mAP improvement on the dataset \cite{d3}  with negation semantics, along with consistent gains on other test sets with positive semantics. These results empirically support our hypothesis that improving negation comprehension also enhances the overall understanding of modifiers.

In summary, our work can be summarized through the following three contributions: 
\begin{enumerate}
    \item We construct \textit{D-Negation}, the first visual grounding dataset with paired positive and negative semantic descriptions across multiple attributes.  
    \item We introduce \textit{Grouped Opposition-Based Learning (GOBL)}, an efficient fine-tuning mechanism that explicitly leverages opposition pairs to strengthen negation comprehension.  
    \item We empirically demonstrate that enhancing negation understanding not only improves performance on negation-related tasks but also boosts general grounding ability on standard benchmarks.  
\end{enumerate}

\begin{figure*}[tbh]
    \centering
    \includegraphics[width=\linewidth]{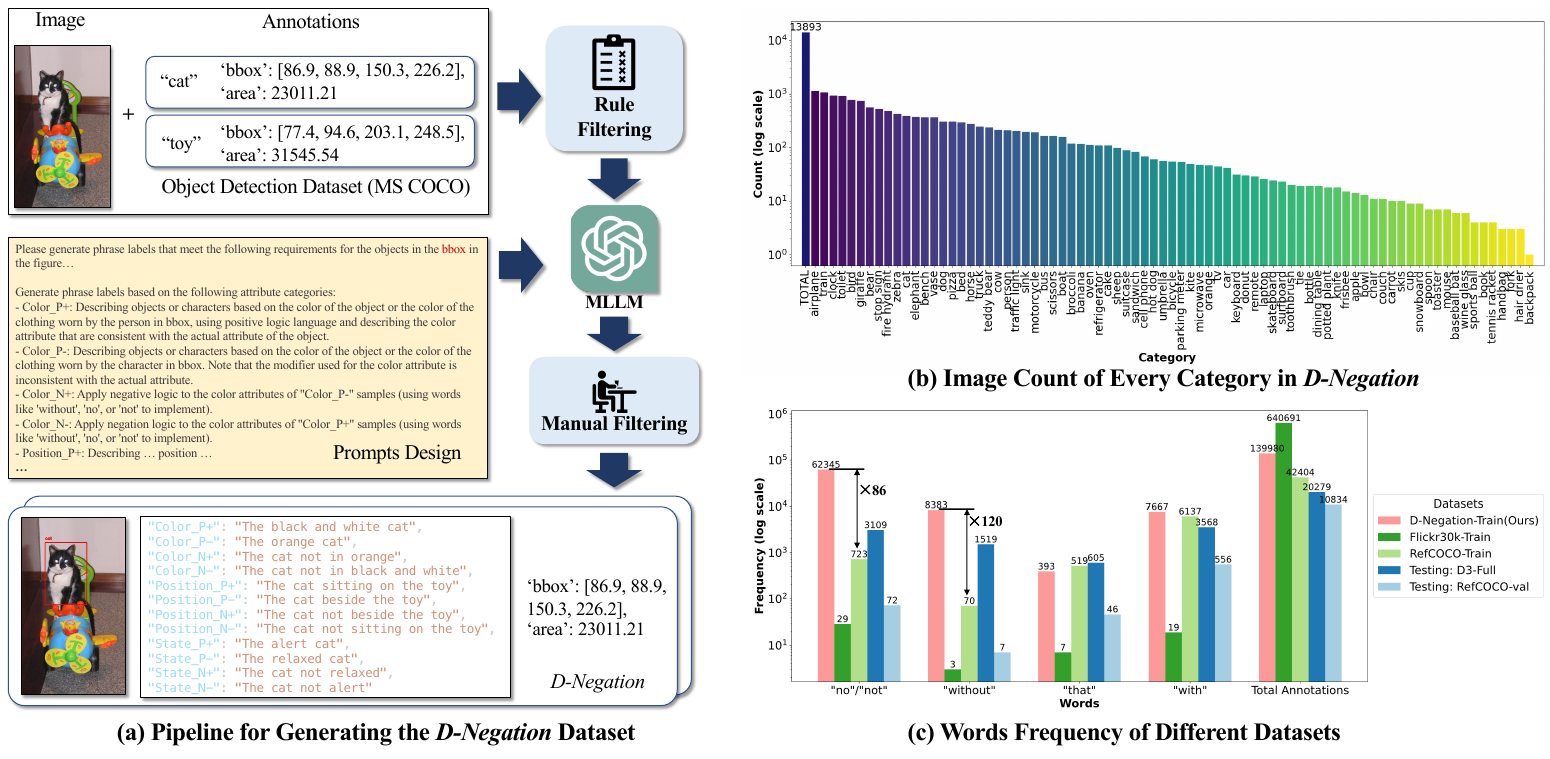}
    \caption{(a) We use the advanced MLLM, GPT-4V, to generate 6 positive semantic and six negative semantic referrings for every selected image. (b) The proposed D-Negation dataset contains a total of 13,893 images across 80 categories. (c) The proposed D-Negation dataset contains 139,980 text annotations. What sets it apart from existing datasets is the significantly higher frequency of negation words, as well as a greater tendency to use modifiers to describe objects.
    }
    \label{fig:framework1}
    \vspace{-0.8em}
\end{figure*}
\section{Related Work}
\label{sec:related}

\subsection{Visual Grounding.}

Visual Grounding (VG)~\cite{Qiao_Deng_Wu_2021, Deng_Yang_Chen_Zhou_Li_2021,yu2016modeling,ad1,ad2} aims to localize image regions corresponding to natural language expressions, serving as a core capability for multimodal understanding and reasoning.
Early approaches often decoupled visual detection and language grounding, limiting their ability to handle complex linguistic structures and diverse visual scenes.

Recent advances have increasingly unified object detection and language grounding within a single framework.
MDETR~\cite{p1}, built upon DETR~\cite{carion2020end}, explicitly aligns visual tokens with linguistic tokens, enabling fine-grained region–word correspondence.
GLIP~\cite{p2} reformulates object detection as a grounding problem through language-aware feature fusion, substantially improving instance-level semantic alignment.
UNINEXT~\cite{p3} introduces a prompt-based formulation that unifies multiple vision–language tasks and enhances generalization to unseen settings.
Grounding DINO~\cite{p4} further strengthens cross-modal interaction by injecting language guidance throughout transformer layers, leading to robust and scalable grounding performance.
APE~\cite{p5} extends this paradigm by proposing an all-in-one alignment and prompting mechanism that bridges object detection and segmentation under a unified grounding framework.

More recent studies have explored complementary directions.
HiVG~\cite{hivg} focuses on hierarchical reasoning to address compositional referring expressions,
MaPPER~\cite{mapper} improves grounding robustness through multi-positive prompt learning,
and Ferret~\cite{ferret} leverages instruction-following and large-scale vision--language pretraining to enhance interactive grounding capabilities.
Despite their strong performance on standard VG benchmarks, these methods primarily target affirmative referring expressions and do not explicitly model negation or exclusion semantics.


\vspace{0.5em}

\subsection{Hard Negatives and Negative-Semantics.}

Hard negatives have long been recognized as a valuable tool in machine learning for enhancing model discriminability and robustness~\cite{hn4,tnnr1,tnnr2}. In the context of vision-language grounding, hard negatives are particularly useful for testing and improving a model’s ability to distinguish between visually and semantically similar entities. For example, contrasting phrases like ``\textit{the red ball in the corner}'' versus ``\textit{the blue ball in the center}'' forces the model to consider fine-grained visual attributes such as color, position, and context. Recent works such as CREPE~\cite{hn1} and NegCLIP~\cite{hn2} incorporate hard negatives during training to emphasize contextual and relational cues, leading to stronger representations and better generalization in grounding tasks.

In parallel, negative semantics introduces prompts that explicitly include negation constructs such as ``not'', ``no'', or ``without''. These prompts are essential for evaluating whether models can correctly understand the \textit{absence} of a concept, rather than its presence. For instance, the phrase ``\textit{the person not wearing a hat}'' requires the model to reason about the negated condition, which is cognitively more challenging. Methods such as CLIPN~\cite{hn3} utilize such constructs to improve out-of-distribution detection by focusing on what is deliberately excluded from the scene. CoN-CLIP~\cite{hn4} goes a step further by using large language models to automatically generate a rich set of negative prompts, forming the CC-Neg dataset and demonstrating notable improvements in image classification.

However, these approaches remain largely confined to classification or coarse-grained recognition tasks. Few methods have effectively integrated hard negatives and negative semantics into the spatially grounded, instance-level setting of visual grounding. Our work addresses this gap by embedding both types of negative signals directly into the grounding process. By jointly modeling what is present and what is explicitly negated, our framework significantly enhances the model's discrimination capacity, particularly in cluttered scenes or under ambiguous linguistic expressions. This integration enables more precise localization and deeper semantic alignment between language and vision.

\vspace{0.5em}

\subsection{Opposition-Based Learning.}

Opposition-Based Learning (OBL)~\cite{ob1} is a principled learning framework that leverages the inherent oppositional relationships between concepts to improve learning efficiency. Inspired by early cognitive theories and optimization strategies~\cite{tizhoosh2005opposition}, OBL differs from conventional contrastive learning~\cite{schroff2015facenet} by enforcing stricter semantic oppositions rather than loosely defined negative samples. In contrastive learning, negative examples are often semantically distant or arbitrarily selected, which may dilute the signal necessary for high-fidelity discrimination. OBL, by contrast, pairs each training sample with a carefully crafted \textit{opposite} example, thereby intensifying the learning signal and accelerating convergence.

While OBL has shown promise in various machine learning domains~\cite{ob3, ob4, ob5}, its application to vision-language tasks remains underexplored—particularly in settings requiring spatial reasoning, such as detection and localization. This is a missed opportunity, as many grounding scenarios naturally lend themselves to opposition-based structures, e.g., ``\textit{the cat on the left}'' versus ``\textit{the cat on the right}'', or ``\textit{the person wearing a mask}'' versus ``\textit{the person not wearing a mask}''.

To address this gap, we propose extending the OBL framework to incorporate both positive and negative semantics within visual grounding models. Specifically, we construct semantically opposed prompt pairs—composed of affirmations and their negations—and use them to guide the learning process. By forming such oppositional pairs, the model is encouraged to learn not just what a phrase describes, but also what it explicitly excludes. This dual learning signal leads to richer, more nuanced representations and enables the model to handle a broader range of natural language inputs with higher precision.

Our proposed framework represents the first systematic effort to integrate opposition-based learning with multimodal grounding, paving the way for more cognitively inspired and semantically aware vision-language systems.

\begin{figure*}[tbh]
    \centering
    \includegraphics[width=\linewidth]{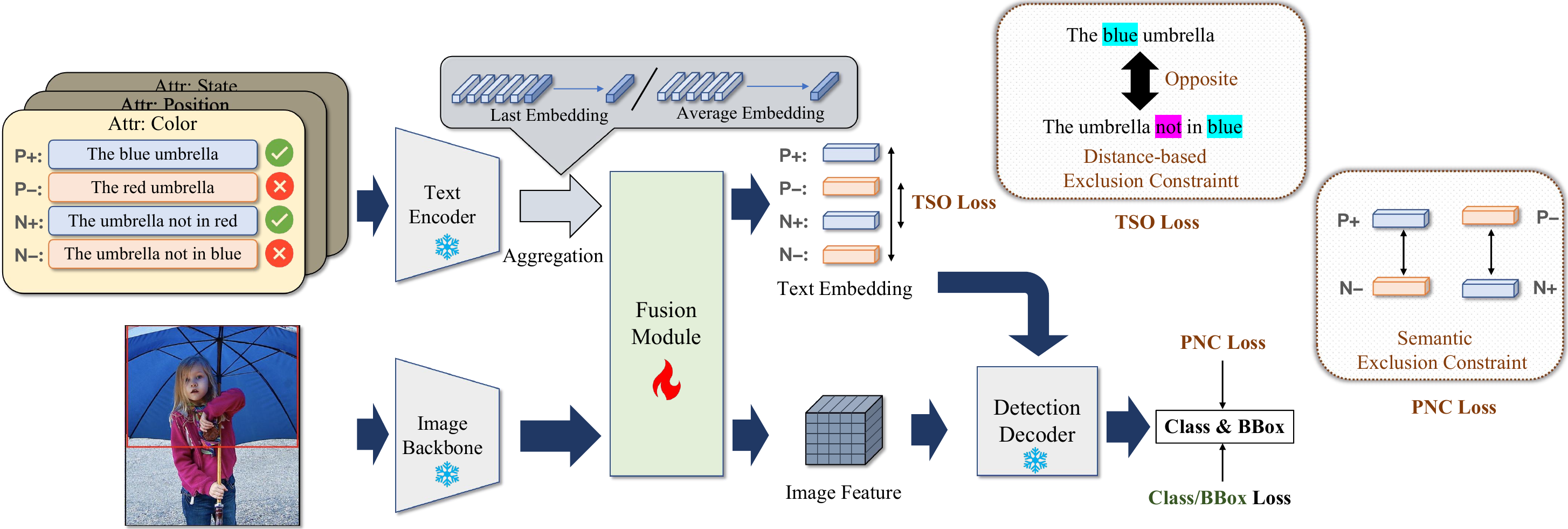}
    \caption{Overview of the proposed fine-tuning framework with explicit semantic-opposition constraints.
    The framework is designed for the \textit{D-Negation} dataset and is compatible with visual grounding models that incorporate a vision--language fusion module.
    Given an image and semantically opposed textual descriptions (e.g., positive vs.\ negative attributes), text embeddings interact with visual features in the fusion module and are decoded into grounding predictions.
    Beyond standard grounding losses, we introduce two complementary objectives.
    The TSO loss imposes a \emph{distance-based exclusion constraint} in the text embedding space, explicitly pushing semantically opposed descriptions (e.g., ``red'' vs.\ ``not red'') apart.
    The PNC loss enforces a \emph{semantic exclusion constraint}, ensuring that a visual region cannot be simultaneously aligned with both polarities of the same attribute.
    Together, these constraints address negation-induced grounding ambiguity by enforcing semantic opposition at both the linguistic and cross-modal fusion levels.}
    \label{fig:framework}
    \vspace{-0.8em}
\end{figure*}

\section{Methodology}
\label{sec:method}

In this section, we present our method in four parts.  
First, we introduce the preliminaries of vision grounding models.  
Second, we describe the prompting strategy used in constructing the \textit{D-Negation} dataset.  
Third, we explain the acquisition process of the \textit{D-Negation} dataset.  
Finally, we detail the efficient fine-tuning mechanism that leverages the constructed dataset.

\subsection{Preliminaries of Vision Grounding Models}

A typical vision grounding model, like GLIP~\cite{p2}, Grounding-DINO~\cite{p4}, and APE~\cite{p5}, consists of four key components: image encoder, language encoder, fusion module, and detection decoder. These components align visual and textual information to perform tasks such as object detection and grounding.

The image is first processed by a visual encoder \(\text{Enc}_I\), utilizing either a Convolutional Neural Network (CNN) \cite{ba1} or a Transformer \cite{ba2, ba3} as its backbone, to extract regional or box features \(O\). Concurrently, every textual \textit{prompt} is processed by a pre-trained language encoder \(\text{Enc}_L\) to generate token features and then aggregate to the text feature \(P\):
\begin{align}
O &= \text{Enc}_I(\text{\textit{Img}}), 
P = \text{Enc}_L(\text{\textit{Prompt}}),
\label{IL1}
\end{align}
A fusion module then processes the features:
\begin{align}
f_q, f_l = \text{Fusion}(O, P).  
\label{IL9}
\end{align}
This module optimizes the synchronization of visual and linguistic elements, enhancing interpretive capabilities. The variables \(f_q\) and \(f_l\) represent the fused feature vectors from the visual and textual data, respectively.


The alignment scores \(S_{\text{cls}}\) between image regions and corresponding words in the prompt are computed as:
\begin{equation}
S_{\text{cls}} =f_l^\top f_q 
\label{IL2}.
\end{equation}
These scores are used to train the model by minimizing the combined classification and localization loss:
\begin{equation}
L = L_{\text{cls}} + L_{\text{loc}}.
\label{IL3}
\end{equation}
\(L_{\text{cls}}\) focuses on aligning visual regions with textual classes using methods like many-to-one or Hungarian matching \cite{hu1,hu2}, often optimized with focal loss. \(L_{\text{loc}}\) ensures accurate bounding box predictions.

The aforementioned content provides an overview of the fundamental architecture of a vision-language grounding model, which typically fuses the visual and linguistic features only at the final stage to compute alignment scores.  In practical implementations, different implementations enhance integration through various detection decoders and vision-language fusion modules.

\subsection{Prompting Strategy}

The primary objective is to obtain high-quality hard negative and negative-semantics samples. Manual annotation is both costly and time-consuming; therefore, leveraging the capabilities of advanced vision-language models, we employ GPT-4V as an annotation tool. By designing a carefully crafted set of prompts, we guide GPT-4V to generate labels that meet our requirements. The detailed workflow is described below.

We first filter out easily annotatable objects in the COCO dataset using a simple rule. Since current MLLMs tend to become confused and hallucinate when annotating multiple objects, we restrict our selection to single-annotated objects, as shown in Algorithm \ref{Rule}. This also mitigates issues with objects that are too small and reduces the risk of generating inaccurate labels. After filtering, we visualize the bounding boxes on the images and provide them, along with the designed instructions, as input to the MLLM for annotation.

\begin{algorithm}
\caption{Rule Filtering} \label{Rule}
\begin{algorithmic}
\FOR{each \textit{image, anns} in \textit{coco\_annotations}}
    \IF{the number of annotations for this \textit{image} is equal to 1}
        \STATE \textit{ann} $\leftarrow$ \textit{anns[0]}
        \STATE \textit{annoted\_image} $\leftarrow$ visualize(\textit{ann}, \textit{image})
        \STATE \textbf{input} \textit{annoted\_image}, \textit{Instruction} \textbf{to} \textit{GPT-4V}
        \STATE \textit{new\_ann} $\leftarrow$ \textbf{output} \textit{from} \textit{GPT-4V} 
        \STATE Add (\textit{image}, \textit{new\_ann}) to dataset
    \ENDIF
\ENDFOR
\end{algorithmic}
\end{algorithm}

After several trials, we found that providing GPT-4V with a template in a strict dictionary format yields the best results. This approach helps in generating phrase labels that are systematically structured, improving consistency and accuracy in attribute description.
In addition, the prompting instructions and formatting details are provided in the supplementary material.

\subsection{Acquisition of the \textit{D-Negation} Dataset}

To systematically capture both positive and negative semantics, we employ GPT-4V with carefully designed prompts to generate diverse object descriptions, including true and hard-negative samples. The complete dataset generation process is outlined below.

The overall workflow is illustrated in Figure~\ref{fig:framework1}. Specifically, Figure~\ref{fig:framework1}(a) shows the data filtering process and the design of prompts. Figure~\ref{fig:framework1}(b) presents the image count for each category in \textit{D-Negation}, and Figure~\ref{fig:framework1}(c) depicts the frequency of negation words across different datasets.

We selected three commonly occurring attributes—``color'', ``position'', and ``state'' as object descriptors. Appropriate prompts were designed to enable GPT-4V to generate four distinct labels for each object:

\begin{itemize}
    \item \textbf{True Description with Positive Semantics (P+)}: A description using positive logic that accurately reflects the object's actual state.
    \item \textbf{False Description with Positive Semantics (P-)}: A description using positive logic that inaccurately reflects the object's actual state.
    \item \textbf{True Description with Negative Semantics (N+)}: A description using negative logic that accurately reflects the object's actual state.
    \item \textbf{False Description with Negative Semantics (N-)}: A description using negative logic that inaccurately reflects the object's actual state.
\end{itemize}

In this context, \textit{Positive-Semantic} refers to affirmative logic descriptions, while \textit{Negative-Semantic} refers to negated logic descriptions. The symbols ``+'' and ``-'' denote whether the description is true or false. For each object, considering the three attributes, a total of 12 describes are generated.

The generated describes exhibit strong interrelationships while remaining independent, each serving a distinct purpose. `P+', described with positive-semantic language, enhances the model’s understanding of specific attributes. `P-', functioning as a hard negative for P+, strengthens the model’s attribute differentiation capabilities. `N+,' described with negative-semantic language, is logically opposed to `P-,' thereby reinforcing the model’s comprehension of negated logic. Similarly, `N-,' serving as a hard negative for `N+,' enhances both the model’s differentiation ability and understanding of negative logic.
The rich semantics and balanced distribution of these four sample types significantly contribute to improving the model's learning efficiency.


\subsection{Efficient Fine-tuning with \textit{GOBL}}

In the previous discussion of \textit{D-Negation}, it was established that a key feature of the dataset is the presence of corresponding labels with negation semantic logic, which provides additional information that can be leveraged by the model.  
We posit that the grounding model's limited sensitivity to negative semantics stems not from the text encoder or detection decoder, but from the fusion module. While the pre-trained text encoder has already been exposed to negative semantic texts, and the decoder can effectively detect objects with positive semantic references, the fusion module often confuses positive and negative features.  
This observation suggests that improving the model's comprehension of negation semantics requires targeted intervention at the fusion stage.

Although the model can be trained using only the standard loss (Eq.~\ref{IL3}), this approach fails to fully exploit the logical relationships between positive and negative descriptions. To address this limitation, we propose two additional grouped loss functions designed to enhance model performance. Both loss functions operate on a triplet consisting of an image, a correct description, and an incorrect description. As illustrated in Figure~\ref{fig:framework}, we group up to 12 semantic annotations per image from the \textit{D-Negation} dataset, forming six pairs of opposing references. Descriptions sharing the same attribute are paired as opposites: `P+' with `N-' and `P-' with `N+'. Within these six groups, `P+' and `N+' are treated as positive samples, while `P-' and `N-' are treated as negative samples. This structured pairing enables the model to explicitly learn the contrasts between positive and negative semantics, addressing the fusion module’s confusion and improving overall grounding performance.

\subsubsection{Positive-Negation Constraint (PNC) Loss}

The Positive-Negative Constraint (PNC) loss is introduced to capitalize on the semantic opposition between positive and negative prompts. First, similarity scores are calculated as follows:
\begin{equation}
s_i = \frac{\langle f_q, f_{t_i} \rangle}{\|f_q\| \cdot \|f_{t_i}\|}, \quad i \in \{P, N\},
\end{equation}
where \(f_q\) represents processed region features, and \(f_{t_i}\) denotes text features after image-language fusion. The normalized similarity score for each prompt is computed as:
\begin{equation}
\bar{S}_{\text{cls}} = \frac{e^{\sigma s_1}}{e^{\sigma s_1} + e^{\sigma s_2}}.
\end{equation}
Here, \(\sigma\) is a control factor that adjusts sensitivity to semantic variations. Finally, the loss function is defined as:
\begin{equation}
L_{\text{\textit{PNC}}} = \text{loss}(\bar{S}_{\text{cls}}; T).
\end{equation}
In this equation, \(T\) represents the target matching between regions and the target prompt, computed using methods such as many-to-one or bipartite Hungarian matching \cite{hu1,hu2}. The PNC loss compels the model to differentiate between opposing prompts, thereby enhancing its understanding of positive and negative logic.

\subsubsection{Text Semantic-Opposite (TSO) Loss}

Previous work on CLIPN \cite{hn3} has shown that models often struggle with positive-negative logic due to the high similarity in feature vectors extracted from prompts \(P\) and \(N\). Inspired by this, we introduce the Text Semantic-Opposite (TSO) loss, which ensures that feature vectors corresponding to opposing prompts are positioned distantly in the feature space. The TSO loss is defined as:
\begin{equation}
L_{\text{\textit{TSO}}} = \frac{1}{N} \left(2 - \sum_{i=1}^{N} \| f_p - f_n \|^2 \right),
\end{equation}
where \(\|\cdot\|_2\) denotes the L2 distance function, and \(f_p\) and \(f_n\) represent the feature vectors of the positive and negative-semantics prompts, respectively. This loss reinforces the model's ability to distinguish between semantically opposite prompts.
The overall loss function is defined as:
\begin{equation}
L_{\text{total}} = L_{\text{cls}} + L_{\text{loc}} + \alpha L_{\text{\textit{PNC}}} + \beta L_{\text{\textit{TSO}}},
\label{ILt}
\end{equation}
where \(\alpha\) and \(\beta\) are weighting factors,  \(L_{\text{cls}}\) and \(L_{\text{loc}}\) correspond to the standard loss components defined in (\ref{IL3}). This integrated loss formulation not only improves the model's ability to distinguish between positive and negative-semantics but also preserves its effectiveness in classification and localization tasks.

\section{Experiments}
\label{sec:expe}

\subsection{Experimental Setup and Training Details}

All experiments related to APE were conducted on a single NVIDIA A6000 GPU, while experiments involving Grounding-DINO were performed on a single NVIDIA V100 GPU. The specific parameter settings employed were \(\sigma = 5\), \(\alpha = 0.5\), and \(\beta = 0.3\). The weights for \(L_{\text{cls}}\), \(L_{\text{loc}}\), and \(L_{\text{giou}}\) were set to 1, 5, and 2, respectively.

\begin{table}[h!]
\caption{Comparison of Training Configurations}
\centering
\footnotesize
\begin{tabular}{lccc}
\toprule
\textbf{Method} & \textbf{Image Consumption} & \textbf{Batch Size} & \textbf{Epochs} \\
\midrule
Grounding-DINO & $\sim$6.8M & 64 & 18 \\
APE & 17.28M & 16 & 1 \\
Ours & 13K & 1 & 1 \\
\bottomrule
\end{tabular}
\label{training_config}
\end{table}

As illustrated in Table~\ref{training_config}, our proposed method achieves a substantial reduction in both data consumption and computational overhead compared to the original training paradigms. Specifically, our method requires only 13K training images—several orders of magnitude fewer than Grounding-DINO ($\sim$6.8M) and APE (17.28M). 

All models using our approach are trained for a single epoch with a batch size of 1. The typical training duration per epoch is approximately 10 hours. Notably, even for the largest model (APE-D), which contains the most parameters among the compared models, training completes within 14 hours per epoch. This is considerably more efficient than the full-scale training procedure originally adopted by APE, which involves processing over 17 million images.

These results underscore the efficiency of our method, which achieves competitive—and in many cases superior—performance while operating under significantly reduced computational and data requirements. This highlights the practical value of our framework in scenarios with limited resources or where rapid adaptation is essential.

\subsection{Combination with State-of-the-art methods}
\subsubsection{Methods and Procedure}
In recent years, numerous methods have exhibited strong performance. To assess the effectiveness of our approach, we selected two state-of-the-art methods, Grounding-DINO \cite{p4} and APE \cite{p5}, as baselines. For APE, we consider four different settings: 
APE-A, based on DETA \cite{ex1} and ViTL \cite{ex2}, is trained solely on detection and segmentation data, including COCO, LVIS, Objects365, OpenImages, and Visual Genome. APE-B is enhanced with Visual Genome region descriptions and RefCOCO \cite{da6}. APE-C incorporates class-agnostic data from SA-1B \cite{da5} for training. APE-D further integrates GQA \cite{hudson2019gqa}, PhraseCut \cite{da4}, and \textit{Flickr30k }\cite{da3}. We utilized the official APE code and released checkpoints as our starting point. Since the training code for Grounding-DINO is unavailable, we opted to use MM-Grounding-DINO \cite{zhao2024open} as an alternative. `Grounding-DINO-Base' refers to the MM-Grounding-DINO-T model trained on the O365 and GoldG datasets~\cite{p2}, while `Grounding-DINO-Full' refers to the MM-Grounding-DINO-T model trained on the full training set (O365, GoldG, GRIT, V3Det).
The majority of hyperparameters were retained from the original methods, with any exceptions documented in the supplementary materials.

\subsubsection{Evaluation Benchmarks}

To the best of our knowledge, the  description detection dataset ($D^3$) \cite{d3} is the only resource specifically designed for evaluating grounding models on negative-semantics prompts. Consequently, $D^3$ was selected as our primary evaluation benchmark. Additionally, we constructed a test set within \textit{D-Negation} to serve as a supplementary benchmark.
In the $D^3$ dataset \cite{d3}, we use the terms \textit{Full} , \textit{Presence}, and  \textit{Absence} to denote evaluations on all descriptions, presence descriptions only, and absence descriptions only, respectively. The \textit{Absence} metric is particularly indicative of a model's capability to detect negative-semantics prompts. The evaluation metrics for $D^3$ include both intra-scenario and inter-scenario evaluations, with detailed configurations available in the DOD \cite{d3} study.
Intra-scenario evaluation is more commonly utilized in typical scenarios and, therefore, serves as the primary performance indicator in our experiments.
For the \textit{D-Negation} dataset, we split 2,000 images to form the test set with uniform distribution. The evaluation procedures applied to \textit{D-Negation} are identical to those used for the $D^3$, which uses standard mAP for evaluation. 
Besides, we use RefCOCO~\cite{kazemzadeh2014referitgame} as an additional positive-semantics test benchmark.

\subsection{Main Results}


\begin{table*}[tbh]
\centering
\caption{Performance comparison on $D^3$ dataset.}
\begin{tabular}{c|ccc|ccc}
\toprule\toprule
\multirow{2}{*}{\centering\textbf{Method}} &\multicolumn{3}{c|}{\textbf{Intra-scenario}} & \multicolumn{3}{c}{\textbf{Inter-scenario}} \\
\cmidrule(r){2-7}
 & \textit{\textbf{Full}}& \textit{\textbf{Presence}} & \textit{\textbf{Absence}} & \textit{\textbf{Full}} & \textit{\textbf{Presence}} & \textit{\textbf{Absence}} \\ \midrule
SPHINX-7B \cite{sphinx} & 10.6 & 11.4 & 7.9 & - & - & - \\ 
GLIP-T ~\cite{p2} & 19.1 & 18.3 & 21.5 & - & - & - \\ 
OFA-DOD \cite{d3} & 21.6 & 23.7 & 15.4 & 5.7 & 6.9 & 2.3 \\ 
FIBER-B ~\cite{fiber} & 22.7 & 21.5 & 26.0 & - & - & - \\
\midrule
MLLM with a larger parameter scale \\
InternVL2-8B ~\cite{inter1} & 9.8 & 11.0 & 6.2 & - & - & - \\
InternVL2-76B ~\cite{internvl2} & 25.3 & 25.7 & 23.5 & - & - & - \\
Griffon(13B) ~\cite{griffonv1} & 12.3 & 12.4 & 12.2 & - & - & - \\
Groma(7B)~\cite{groma} & 16.0 & 15.9 & 16.3 & - & - & - \\

\midrule
Grounding-DINO-Base & 15.6 & 16.4 & 13.4 & - & - & - \\
Grounding-DINO-Full & 17.5 & 18.0 & 15.9 & - & - & - \\
\rowcolor{mygray} Grounding-DINO-Base(+Ours) & 17.8 \newline \textit{(+2.2)} & 17.4 \newline \textit{(+1.0)} & 19.0 \newline \textbf{\textit{(+5.6)}} & - & - & - \\
\rowcolor{mygray} Grounding-DINO-Full(+Ours) & 19.0 \newline \textit{(+1.5)} & 18.5 \newline \textit{(+0.5)} & 20.9 \newline \textbf{\textit{(+5.0)}} & - & - & - \\
\midrule
APE-A & 25.1 & 24.5 & 26.9 & 16.4 & 15.9 & 17.9 \\
APE-B & 30.0 & 29.9 & 30.3 & 20.0 & 20.5 & 18.6 \\
APE-C & 27.8 & 27.9 & 27.3 & 20.4 & 21.2 & 18.1 \\
APE-D & 37.5 & 38.8 & 33.9 & 21.0 & 22.0 & 17.9 \\
\rowcolor{mygray} APE-A(+Ours)  & 27.1 \newline \textit{(+2.0)} & 26.7 \newline \textit{(+2.2)} & 27.8 \newline \textit{(+0.9)} & 17.8 \newline \textit{(\textbf{+1.4})} & 17.2 \newline \textit{(\textbf{+1.3})} & 18.8 \newline \textit{(+0.9)} \\
\rowcolor{mygray} APE-B(+Ours)  & 32.4 \newline \textit{(+2.4)} & 32.2 \newline \textit{(+2.3)} & 32.9 \newline \textit{(+2.6)} & 20.9 \newline \textit{(+0.9)} & 21.0 \newline \textit{(+0.5)} & 19.0 \newline \textit{(+0.4)} \\
\rowcolor{mygray} APE-C(+Ours) & 32.5 \newline \textit{(\textbf{+4.7})} & 32.3 \newline \textit{(\textbf{+4.4})} & 33.0 \newline \textit{\textbf{(+5.7})} &21.2 \newline \textit{(+0.8)} &21.8\newline \textit{(+0.6)} & \textbf{19.3} \newline \textit{(\textbf{+1.2})}\\
\rowcolor{mygray} APE-D(+Ours) & \textbf{38.6} \newline \textit{(+1.1)} & \textbf{39.8} \newline \textit{(+1.0)} & \textbf{35.0} \newline \textit{(+1.1)} & \textbf{22.4} \newline \textit{(\textbf{+1.4})} & \textbf{23.2} \newline \textit{(+1.2)} & 19.0 \newline \textit{(+1.1)} \\
\bottomrule\bottomrule
\end{tabular}
\label{main1}
\end{table*}

\begin{table}[h!]
\centering
\caption{Performance comparison on \textit{D-Negation} test set. }
\resizebox{\linewidth}{!}{
\setlength\tabcolsep{3pt}
\begin{tabular}{c|cc>{\columncolor{mygray}}c}
\toprule\toprule
\textbf{Grounding Models} & \textit{\textbf{Original}} & \textit{\textbf{+Flickr30k}} & \textbf{+Ours} \\
\midrule
Grounding-DINO-Base & 71.1 & 72.6 \newline \textit{(+1.5)} & \textbf{75.3 \newline \textit{(+4.2)} }\\
Grounding-DINO-Full& 74.2 & 74.0 \newline \textit{(-0.2)} & \textbf{77.0 \newline \textit{(+2.8)} }\\
APE-A & 76.8 & 75.0 \newline \textit{(-1.8)} & \textbf{78.9 \newline \textit{(+2.1)} }\\
APE-B & 80.5 & 78.9 \newline \textit{(-1.6)} & \textbf{83.7 \newline \textit{(+3.2)} }\\
APE-C & 78.6 & 80.1 \newline \textit{(+1.4)} & \textbf{82.8 \newline \textit{(+4.2)} }\\
APE-D & 78.9 & 80.2 \newline \textit{(+1.3)} & \textbf{84.1 \newline \textit{(+5.2)} }\\
\bottomrule\bottomrule
\end{tabular}
}
\label{main2}
\end{table}


\begin{table*}[h]
\centering
\caption{Performance comparison of different methods on RefCOCO val, testA, and testB datasets.}
\begin{tabular}{c|ccc|ccc|ccc}
\toprule\toprule
\multirow{2}{*}{\centering\textbf{Method}}& \multicolumn{3}{c|}{\textbf{val}} & \multicolumn{3}{c|}{\textbf{testA}} & \multicolumn{3}{c}{\textbf{testB}} \\
                & \textbf{@1} & \textbf{@5} & \textbf{@10} & \textbf{@1} & \textbf{@5} & \textbf{@10} & \textbf{@1} & \textbf{@5} & \textbf{@10} \\
\midrule
Grounding-DINO-Full & 53.1 & 89.7 & 95.1 & 59.2 & 91.0 & 95.5 & 46.7 & 87.8 & 93.6 \\
\rowcolor{mygray} Grounding-DINO-Full(+Ours) & 51.6 & 87.2 & 93.3 & \textbf{60.3} & \textbf{92.6} & \textbf{96.8} & \textbf{48.4} & \textbf{88.2} & \textbf{95.0} \\
APE-C & 79.8 & - & - & 86.8 & - & - & 76.2 & - & - \\
\rowcolor{mygray} APE-C(+Ours) & \textbf{80.5} & - & - & \textbf{87.8} & - & - & \textbf{77.1} & - & - \\
\bottomrule\bottomrule
\end{tabular}
\label{main3}
\end{table*}

\begin{table*}[h!]
\centering
\begin{minipage}[t]{0.45\textwidth}
\centering
\caption{The impact of positive and negative semantic samples in \textit{D-Negation} }
\resizebox{\linewidth}{!}{
\setlength\tabcolsep{3pt}
\begin{tabular}{l|ccc}
\toprule\toprule
\textbf{Method} & \textit{\textbf{Full}} & \textit{\textbf{Presence}} & \textit{\textbf{Absence}} \\
\midrule
Baseline(APE-C) & 27.8 & 27.9 & 27.3 \\
+Positive Only & 28.1 \newline \textit{(+0.3)} & 28.8 \newline \textit{(+0.9)} & 27.0 \newline \textit{(-0.3)} \\
+Negative Only & 27.4 \newline \textit{(-0.4)} & 27.2 \newline \textit{(-0.7)} & 27.9 \newline \textit{(+0.6)} \\
+Full \textit{D-Negation} & 28.7 \newline \textit{(+0.9)} & 28.5 \newline \textit{(+0.6)} & 29.1 \newline \textit{(+1.8)} \\
+\textit{D-Negation} + \textit{GOBL} & \textbf{32.5 \newline \textit{(+4.7)}} & \textbf{32.3 \newline \textit{(+4.4)}} & \textbf{33.0 \newline \textit{(+5.7)}} \\
\bottomrule\bottomrule
\end{tabular}
}
\label{main4}
\end{minipage}
\hfill
\begin{minipage}[t]{0.54\textwidth}
\centering
\caption{The impact of the proposed loss functions. Performance comparison on Intra-scenario settings of $D^3$ dataset. Baseline: APE-C.}
\resizebox{\linewidth}{!}{
\setlength\tabcolsep{3pt}
\begin{tabular}{ccc|ccc}
\toprule\toprule
\textbf{\textit{D-Negation}} &  \textbf{\textit{TSO} Loss} &  \textbf{\textit{PNC} Loss} & \textit{\textbf{Full}} & \textit{\textbf{Presence}} & \textit{\textbf{Absence}} \\
\midrule
& & & 27.8 & 27.9 & 27.3 \\
\checkmark & &  & 28.7 \newline \textit{(+0.9)} & 28.5 \newline \textit{(+0.6)} & 29.1 \newline \textit{(+1.8)} \\
\checkmark & \checkmark &  & 29.2 \newline \textit{(+1.4)} & 29.1 \newline \textit{(+1.2)} & 29.5 \newline \textit{(+2.2)} \\
\checkmark & & \checkmark & 32.1 \newline \textit{(+4.3)} & 31.0 \newline \textit{(+3.2)} & 32.5 \newline \textit{(+5.2)} \\
\checkmark & \checkmark & \checkmark & \textbf{32.5 \newline \textit{(+4.7)}} & \textbf{32.3 \newline \textit{(+4.4)}} & \textbf{33.0 \newline \textit{(+5.7)}} \\
\bottomrule\bottomrule
\end{tabular}
}
\label{main5}
\end{minipage}
\end{table*}

We summarize the performance gains achieved by our proposed method in Table~\ref{main1}. For the Grounding-DINO framework, our approach consistently improves the results across both model variants, with the most notable gain observed in the \textit{Absence} metric under negative-semantic scenarios, reaching an improvement of +5.6. These results highlight the effectiveness of our method in handling negation-based referring expressions.

Our approach also leads to consistent and substantial improvements across all four APE model variants, with the most significant gains observed in APE-C. Specifically, we observe improvements of +4.7 in the \textit{Full} metric, +4.4 in \textit{Presence}, and +5.7 in \textit{Absence}. These metrics collectively measure model performance under different semantic conditions: full-spectrum evaluation, referring expressions with only positively stated objects, and referring expressions involving object absence, respectively.
Remarkably, even in the \textit{Presence} setting—where the input only contains affirmatively stated referring expressions—our method still yields performance improvements. This suggests that the proposed opposite supervision strategy not only enables the model to comprehend negation more effectively, but also enhances its overall understanding of linguistic modifiers, such as adjectives and qualifiers, leading to a more precise semantic alignment between vision and language.

It is worth noting that the observed improvements on APE-D are relatively modest compared to other variants. This may be attributed to its significantly larger parameter count, which could potentially lead to a saturation effect or require additional tuning strategies to fully leverage the benefits of our method.
Considering both the amount of training data and the computational cost typically required to achieve comparable improvements, our method demonstrates a favorable trade-off between effectiveness and efficiency.

Table~\ref{main2} presents a targeted evaluation of model performance on the Negation Test benchmark. The \textit{Original} column reports the average precision (AP) scores of the baseline models, while the \textit{Flickr30K} column represents models trained with an equivalent quantity of data randomly sampled from the \textit{Flickr30K} dataset. These results reveal a critical insight: simply increasing the volume of training data does not necessarily enhance the model's ability to handle negative semantics; in certain cases, it may even lead to a degradation in performance.

In contrast, our method produces substantial and consistent gains across all evaluated metrics. These improvements clearly demonstrate that our targeted opposite learning strategy is more effective than naïvely scaling up training data, particularly in equipping models with a nuanced understanding of negation and absence-related linguistic constructs.

\begin{figure*}[tbh]
    \centering
    \includegraphics[width=1.0\linewidth]{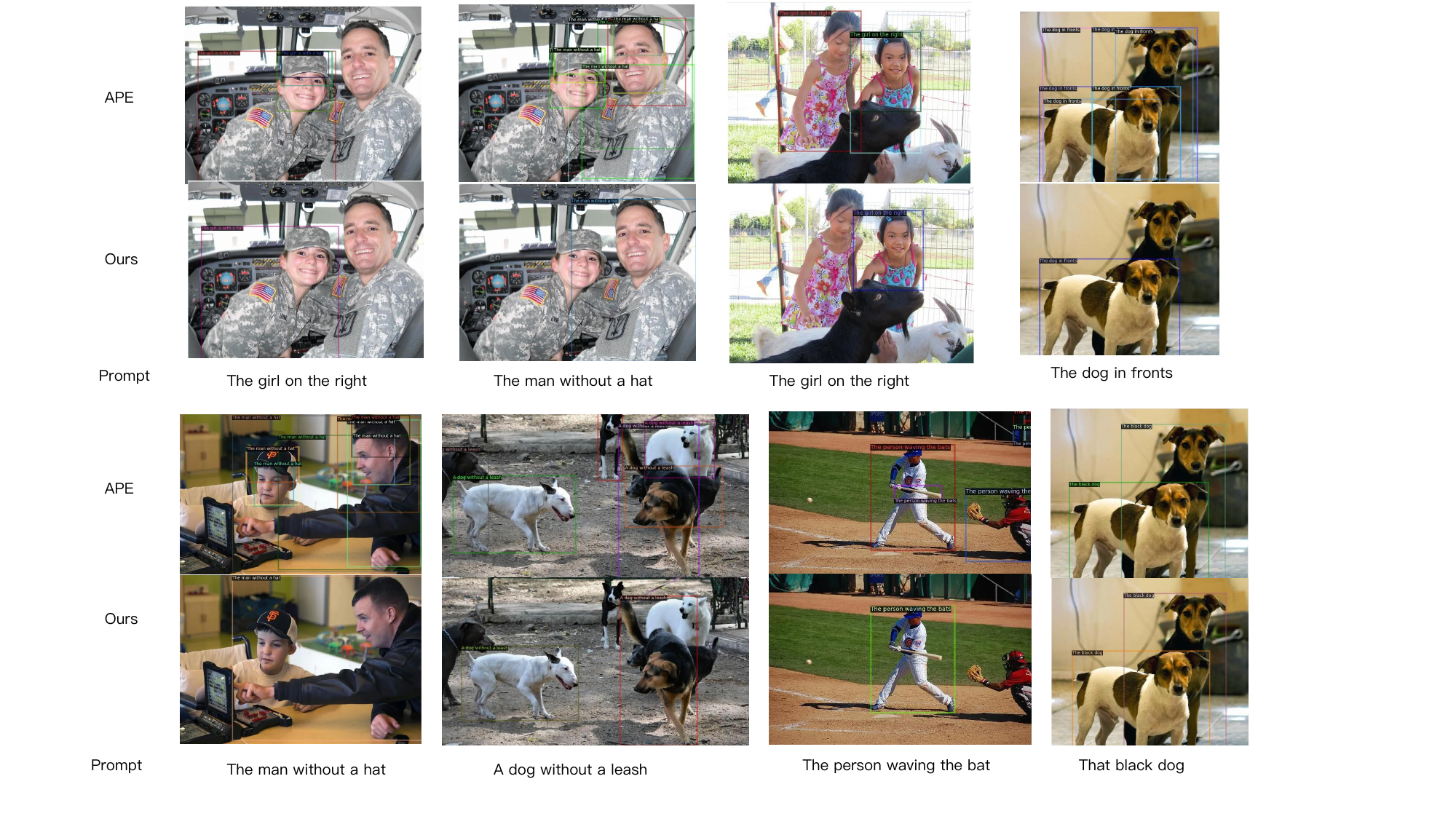}
    \caption{\textbf{Qualitative Comparison of Model Outputs.} Visualization of the APE model's outputs on representative examples after fine-tuning with our proposed method. For clarity, the prompt tokens within bounding boxes are displayed beneath each image.}
    \label{prompt_f1}
    \vspace{-0.8em}
\end{figure*}

\subsection{Domain Generalization}

In this section, we evaluate the ability of our approach to generalize across domains, particularly in scenarios not seen during training. Our goal is to ensure that improvements in handling negative semantics do not come at the cost of reduced robustness in out-of-distribution (OOD) settings.

\subsubsection{Generalization to RefCOCO Variants}

Table~\ref{main3} reports the performance of our method on different splits of the RefCOCO dataset.
To examine the trade-off between semantic precision and generalization, we evaluate two representative models, Grounding-DINO and APE, which exhibit the strongest baseline performance.
Overall, our fine-tuned models achieve consistent, albeit modest, improvements on the RefCOCO test splits, indicating that the proposed semantic alignment does not harm cross-domain generalization.

Notably, our method improves performance on RefCOCO testA and testB by +1.1 and +1.7 at @1, respectively, while exhibiting a small decrease on the val split ($-1.5$ at @1).
We attribute this behavior to the characteristics of RefCOCO, which predominantly contains affirmative referring expressions and offers limited supervision for negation semantics.
Grounding-DINO-Full is fully fine-tuned on large-scale grounding data and is already close to performance saturation in such affirmative-only settings, particularly on the val split that closely matches the training distribution.
In contrast, our method introduces explicit opposition-based supervision to enhance sensitivity to negative semantics.
This targeted supervision acts as a form of structured regularization, which may slightly affect performance in purely affirmative scenarios, while improving generalization on more diverse test distributions.
Importantly, this effect is limited in magnitude and does not impact test performance.
More critically, it enables substantial gains on negation- and absence-oriented benchmarks, which are the primary focus of this work.

\subsubsection{Scalability with Heterogeneous Data Sources}

To evaluate the extensibility of our learning mechanism, we experiment with combining datasets of differing semantic structure. Specifically, Table~\ref{main6} demonstrates the performance of a model trained with a 1:1 mixture of the \textit{Flickr30k} dataset and our proposed \textit{D-Negation}, augmented with the \textit{GOBL} mechanism.

The combination results in the most substantial performance gains across both evaluation settings, affirming the scalability and adaptability of our method. However, this training configuration requires approximately double the computational cost compared to using \textit{D-Negation} alone. As such, while it highlights the potential of our method under larger data regimes, we do not position this setup as our primary contribution.

Across these experiments, we demonstrate that our approach improves semantic robustness without degrading domain generalization. Furthermore, its effectiveness persists across diverse datasets and semantic attribute compositions, underscoring the practicality of our method in real-world multimodal applications.

\subsection{Ablation Experiments}

\begin{table*}[h!]
\centering
\caption{Performance comparison on Intra-scenario of $D^3$ and \textit{D-Negation test} datasets. Our method also performs better when using flick30k as an additional training set.}
\begin{tabular}{l|ccc|c}
\toprule\toprule
\textbf{Method} & \textit{\textbf{Full}} & \textit{\textbf{Presence}} & \textit{\textbf{Absence}} & \textbf{ \textit{D-Negation} } \\
\midrule
Baseline(APE-C) & 27.8 & 27.9 & 27.3 & 78.6 \\
+\textit{\textit{D-Negation}}+\textit{GOBL} & 32.5 \newline \textit{(+4.7)} & 32.3 \newline \textit{(+4.4)} & 33.0 \newline \textit{(+5.7)} & 82.8 \newline \textit{(+3.2)} \\
+Flick30k+\textit{\textit{D-Negation}}+\textit{GOBL} & \textbf{32.9 \newline \textit{(+5.1)}} & \textbf{32.7 \newline \textit{(+4.8)}} & \textbf{33.5 \newline \textit{(+6.2)}} & \textbf{84.1 \newline \textit{(+5.5)}} \\
\bottomrule\bottomrule
\end{tabular}
\label{main6}
\end{table*}

We conducted a comprehensive study in this section to validate the effectiveness of the \textit{D-Negation} dataset and the proposed \textit{GOBL} fine-tuning mechanism. Given the notable impact of our method on the APE-C model, it was chosen as the experimental model for this analysis. We also examined the effects of training our method in conjunction with other datasets.

\subsubsection{Ablation on the \textit{D-Negation} Dataset}
Table~\ref{main4} presents the impact of using different sample types from the \textit{D-Negation} dataset. “+Positive Only” denotes training with only positive semantic samples, while “+Negative Only” refers to training with solely negative semantic samples. Other rows follow similar interpretations.

The results reveal that training with only positive samples brings limited improvement, while using only negative samples even degrades the model's performance. In contrast, combining both types of semantics yields the most substantial performance gain. This clearly demonstrates the necessity of incorporating semantic opposition during training, highlighting the complementarity between positive and negative semantics.

\subsubsection{Ablations of Tuning Module}

To rigorously test this hypothesis, we conduct a controlled comparison of tuning locations under identical training budgets, using APE-C as the base model.
Negation supervision is applied to only one module at a time—namely the text encoder, image backbone, decoder, or fusion module—while all other factors, including training data, loss functions, and optimization schedules, are held constant.
In addition, we perform complementary experiments by fine-tuning different modules of the Grounding-DINO-Base model; detailed results and analyses are provided in the supplementary material.

\begin{table}[t]
\centering
\caption{Relative performance gains (\%) obtained by applying negation supervision to different modules under identical training budgets.}
\label{tuning_module}
\begin{tabular}{lccc}
\toprule
\textbf{Editing Location} & \textbf{Full} & \textbf{Presence} & \textbf{Absence} \\
\midrule
Text Encoder      & +0.7 & +0.5 & +1.1 \\
Image Backbone    & -0.3 & -0.1 & -0.7 \\
Decoder           & +1.2 & +0.6 & +1.3 \\
Fusion Module     & \textbf{+4.7} & \textbf{+4.4} & \textbf{+5.7} \\
\bottomrule
\end{tabular}
\end{table}

As shown in Table~\ref{tuning_module}, tuning the fusion module yields substantially larger gains than all other locations.
Text encoder tuning provides only marginal improvement, indicating that linguistic polarity alone is insufficient without visual grounding.
Image backbone tuning degrades performance, suggesting incompatibility between negation supervision and low-level perceptual learning.
Decoder-level tuning offers limited benefits, consistent with late-stage correction rather than addressing the root cause.
Notably, the fusion module delivers the largest improvements on the absence subset, where exclusion reasoning is critical.
These results demonstrate that negation-related grounding errors primarily emerge during cross-modal fusion, establishing the fusion module as the key structural bottleneck for learning logical exclusion.

\subsubsection{Ablations of Proposed Loss Functions and Hyperparameter Sensitivity}

Table~\ref{main5} examines the contribution of individual components in the proposed \textit{GOBL} loss.
Training with the \textit{D-Negation} dataset alone already improves grounding performance, indicating the benefit of explicit negation supervision.
Introducing the proposed \textit{TSO} and \textit{PNC} losses yields further substantial gains.
In particular, the \textit{PNC} loss alone provides a significant improvement of +4.3, highlighting its strong alignment with the structured semantic opposition encoded in \textit{D-Negation}.
These results suggest that effective negation grounding requires not only contrastive alignment, but also explicit structural constraints on semantic exclusion.

We further analyze the sensitivity of the proposed method to key hyperparameters $\sigma$, $\alpha$, and $\beta$.
As summarized in Table~\ref{hyperparam}, performance remains stable across a broad range of values.
The default configuration ($\sigma=5$, $\alpha=0.5$, $\beta=0.3$) consistently achieves the best or near-best results, while deviations from these settings result in only minor performance fluctuations.
Notably, the absence subset exhibits slightly higher sensitivity, which is expected given its stronger dependence on precise semantic opposition. In addition, we report the corresponding ablation results on APE-B in the supplementary material.
Overall, these results demonstrate that the proposed method is robust and does not rely on careful hyperparameter tuning.

\begin{table}[t]
\centering
\caption{Sensitivity analysis of hyperparameters $\sigma$, $\alpha$, and $\beta$.
All experiments are conducted on APE-C with other parameters fixed to default values.}
\label{hyperparam}
\begin{tabular}{c c c c}
\toprule
\textbf{Setting} & \textbf{Full} & \textbf{Presence} & \textbf{Absence} \\
\midrule
$\sigma=3$   & +4.3 & +4.0 & +5.2 \\
$\sigma=5$   & \textbf{+4.7} & \textbf{+4.4} & \textbf{+5.7} \\
$\sigma=7$   & +4.4 & +4.1 & +5.4 \\
\midrule
$\alpha=0.3$ & +4.1 & +3.8 & +5.0 \\
$\alpha=0.5$ & \textbf{+4.7} & \textbf{+4.4} & \textbf{+5.7} \\
$\alpha=0.7$ & +4.2 & +3.9 & +5.1 \\
\midrule
$\beta=0.1$  & +4.3 & +4.1 & +5.3 \\
$\beta=0.3$  & \textbf{+4.7} & \textbf{+4.4} & \textbf{+5.7} \\
$\beta=0.5$  & +4.2 & +3.9 & +5.0 \\
\bottomrule
\end{tabular}
\end{table}

\subsubsection{Attribute Contribution and Transferability}

\begin{figure}[tbh]
    \centering
    \includegraphics[width=\linewidth]{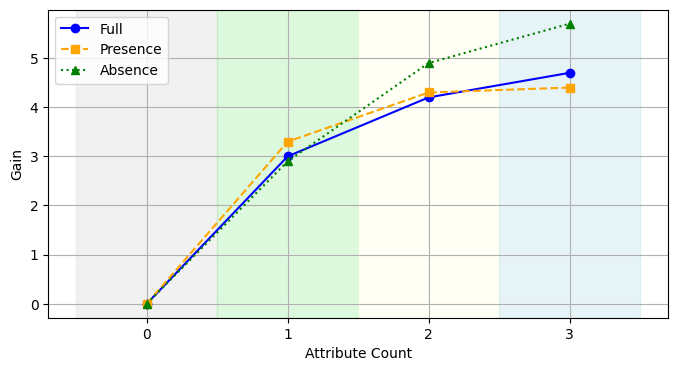}
    \caption{Performance gain of single and combined attributes on the $D^3$ dataset.
    Each attribute enhances the model's performance, and combining multiple attributes leads to even greater improvements. As more attributes are added, the performance gain gradually decreases and eventually stabilizes.
    }
    \label{attri}
    \vspace{-0.8em}
\end{figure}

\begin{figure}[tbh]
    \centering
    \includegraphics[width=\linewidth]{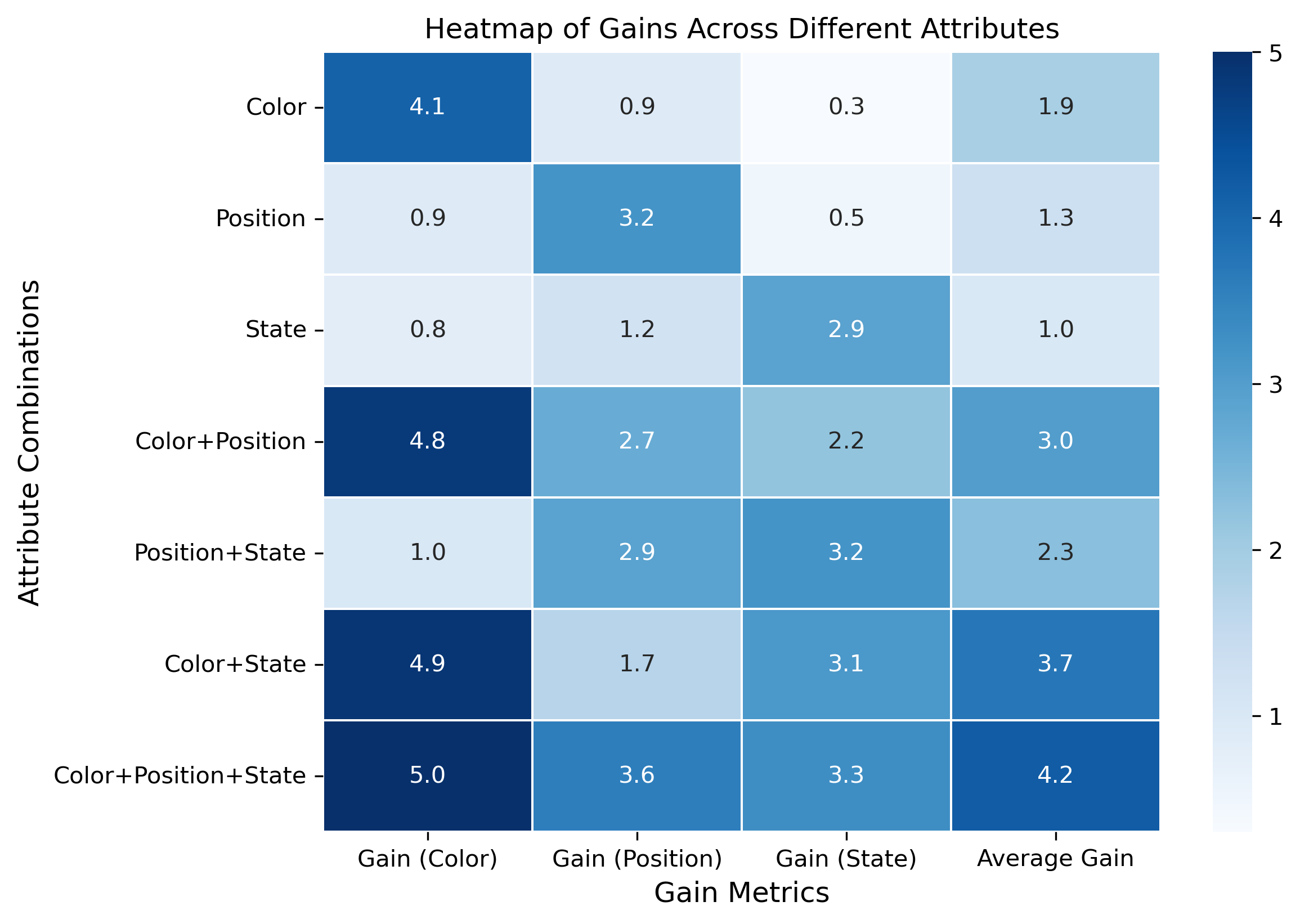}
    \caption{ Heatmap of gains across different attributes. 
    Each attribute enhances performance in its domain and benefits others, demonstrating transferability. The best results occur when all attributes are combined.
    Every attribute significantly boosts performance within its own domain, while also positively influencing the performance in other domains, demonstrating some degree of transferability. The best performance is achieved when all attributes are combined. 
    }
    \label{map}
    \vspace{-0.8em}
\end{figure}

We evaluated the impact of individual attributes and their combinations on model performance by incrementally increasing the number of attributes from 0 to 3 and averaging results across combinations. As shown in Figure \ref{attri}, each attribute enhances performance, with diminishing returns as more attributes are added.
Figure \ref{map} demonstrates the influence of attributes in the \textit{D-Negation}. Each attribute significantly improves performance within its domain while also positively impacting other domains, indicating transferability. The best performance is achieved when all attributes are combined.


\section{Limitations and Future Work}

As illustrated in Figure~\ref{prompt_f1}, fine-tuning on the \textit{D-Negation} dataset substantially improves the model’s ability to reason about attributes and negation semantics; however, several limitations remain.
Due to the limited scale of negation-focused training data, the model still struggles with challenging attribute-level ambiguities, particularly when visual distinctions are subtle (e.g., differentiating a \emph{black} object from a \emph{black-and-white} one), where multiple candidate regions may be incorrectly activated.
Moreover, our current approach primarily enhances negation reasoning at the fusion stage, while leaving the backbone visual representations unchanged, which constrains the model’s ability to resolve fine-grained attribute conflicts under weak or ambiguous visual cues.
Future work will explore extending opposition-based learning to the visual backbone itself, enabling more discriminative attribute representations, and expanding negation-aware training data to further improve robustness in complex attribute and negation scenarios.
\section{Conclusion}
\label{sec:conclusion}
This research introduces innovative methods to enhance grounding models in handling negation by utilizing the \textit{D-Negation} dataset and the \textit{GOBL} mechanism. Our findings confirm that targeted fine-tuning with specialized contrastive loss significantly improves the models' capability to ground complex prompts involving negation.
The \textit{D-Negation} dataset, enriched with discriminative samples embodying both positive and negative semantics, enables vision-language models to overcome traditional training constraints, fostering a deeper understanding of linguistic semantics within visual contexts. The \textit{GOBL} approach prioritizes efficiency, yielding substantial performance gains with minimal modifications to model parameters crucial for scenarios with limited resources or where data acquisition and processing costs are excessive. 
The current limitation lies in the scale and diversity of the dataset. Due to constraints in data sources and the difficulty of annotation, our images rarely contain multiple instances of the same class simultaneously. This necessitates further research to better align with real-world scenarios.

\begin{refcontext}[sorting = none]
\printbibliography
\end{refcontext}

\section{Supplementary Material}

\subsection{Prompting Strategy}

We construct the negation-aware annotation data from the COCO dataset using a rule-based filtering and prompting pipeline.
To reduce annotation ambiguity and hallucination from multi-modal large language models (MLLMs), we first filter out images that contain multiple annotated objects.
Specifically, we retain only images with a single annotated instance, which simplifies object attribution and avoids issues caused by small or overlapping objects.

Given the filtered annotations, we visualize the bounding box on the corresponding image and feed the annotated image into the MLLM together with carefully designed prompting instructions.
This procedure ensures that the model focuses on a single target object with explicit spatial grounding.



Through empirical exploration, we find that enforcing a strict dictionary-style output format leads to more consistent and accurate attribute descriptions.
This structured prompting strategy constrains the generation space of GPT-4V, thereby reducing noise and improving the reliability of the generated phrase labels.
Algorithm~\ref{Prompts} presents the complete prompting instructions and the expected output format used in our experiments.

\begin{algorithm*}
\begin{flushleft}
\caption{Instruction of MLLM} \label{Prompts}
\textit{Please generate phrase labels that meet the following requirements for the objects in the bbox in the figure. Note: The format should strictly be a dictionary where each key corresponds to a specific attribute category, and the values should be appropriate descriptive phrases.}

\textit{1. Format: Use dictionary format to return.}

\textit{2. Example of expected format:}
\begin{itemize}
    \item \textit{Color\_P+: “The man in a red shirt"}
    \item \textit{Color\_P-: “The man in a blue shirt"}
    \item \textit{Color\_N+: “The man not in a blue shirt"}
    \item \textit{Color\_N-: “The man not in a red shirt"}
    \item \textit{Position\_P+: “The man holding the child"}
    \item \textit{Position\_P-: “The man beside the child"}
    \item \textit{Position\_N+: “The man not beside the child"}
    \item \textit{Position\_N-: “The man not holding the child"}
    \item \textit{State\_P+: “The standing man"}
    \item \textit{State\_P-: “The sitting man"}
    \item \textit{State\_N+: “The man not sitting"}
    \item \textit{State\_N-: “The man not standing"}
\end{itemize}

\textit{Now, please generate phrase labels based on the following attribute categories:}
\begin{itemize}
    \item \textit{Color\_P+: Describes objects or characters based on the actual color attribute using positive logic.}
    \item \textit{Color\_P-: Describes objects or characters using a color attribute that is inconsistent with the actual attribute.}
    \item \textit{Color\_N+: Applies negation to the color attributes described in “Color\_P-".}
    \item \textit{Color\_N-: Applies negation to the color attributes described in “Color\_P+".}
    \item \textit{Position\_P+: Describes the position of objects or characters in bbox, ensuring the description matches the actual position attribute.}
    \item \textit{Position\_P-: Describes the position using modifiers that are inconsistent with the actual position attribute.}
    \item \textit{Position\_N+: Applies negation to the position attributes in “Position\_P-".}
    \item \textit{Position\_N-: Applies negation to the position attributes in “Position\_P+".}
    \item \textit{State\_P+: Describes objects or characters based on their actual state using positive logic.}
    \item \textit{State\_P-: Describes the state using modifiers that are inconsistent with the actual state attribute.}
    \item \textit{State\_N+: Applies negation to the state attributes described in “State\_P-".}
    \item \textit{State\_N-: Applies negation to the state attributes described in “State\_P+".}
\end{itemize}
\end{flushleft}
\end{algorithm*}

This structured prompting strategy facilitates the generation of precise and contextually relevant descriptions, enhancing the model's understanding and interpretative capabilities within specified scenarios.
\subsection{Ablations of Tuning Module}

\begin{table}[h!]
\centering
\caption{Comparison of different tuning modules. Base on Grounding-DINO-Base. Performance on Intra-scenario of $D^3$ dataset.}
\resizebox{\columnwidth}{!}{%
\setlength\tabcolsep{3pt}
\begin{tabular}{lccc}
\toprule\toprule
\textbf{Tuning Module} & \textbf{Params(M)} & \textbf{Proportion(\%)} & \textit{\textbf{Full/Presence/Absence}}\\ 
\midrule
Fusion Only    & \textbf{15.7}  & \textbf{9.0}  & \textbf{17.8 / 17.4 / 19.0} \\

+Decoder       & 26.7  & 15.5 &  15.4 / 15.9 / 16.9 \\

+Decoder +Neck & 36.6  & 21.1  &  13.4 / 13.6 / 12.9 \\ 
\bottomrule\bottomrule
\end{tabular}%
}
\label{main7}
\end{table}

Table~\ref{main7} compares the effect of fine-tuning different modules in Grounding-DINO-Base.
We observe that tuning additional components does not necessarily improve performance and can even be detrimental, likely due to interference with pre-trained representations.
The best results are obtained when only the fusion module is tuned, providing initial evidence that semantic polarity errors mainly arise during vision--language interaction.

\subsection{Comparative Analysis of Visual-Language Models}

The \textit{D-Negation} dataset is designed based on the ways in which humans perceive and comprehend entities in the world. In this framework, GPT-4V functions as a tool for the bulk generation of labels that meet specific criteria. To mitigate concerns regarding potential over-reliance on GPT-4V within the dataset, we also explore the performance of other visual-language models through comparative analysis. The methodology employed involves utilizing the same image and prompt across different visual-language models to generate labels, thus enabling a systematic comparison of their respective outputs.

\subsection{Comparison Between GOBL and Standard Contrastive Learning}

To evaluate the effectiveness of the proposed Efficient Fine-tuning strategy with the GOBL mechanism, we conduct a comparative study against standard contrastive-learning-based fine-tuning.
We adopt APE-B and APE-C as baseline models and apply two fine-tuning strategies: 
(i) the original contrastive learning objective used in prior work, and 
(ii) our proposed GOBL-based fine-tuning.

All models are trained under identical settings, including training data, optimization schedules, and evaluation protocols.
Performance is reported on the intra-scenario split of the $D^3$ dataset, with results summarized in Table~\ref{main8}.

\begin{table}[t]
    \centering
    \caption{Performance comparison on the intra-scenario split of the $D^3$ dataset.
    Numbers in parentheses denote absolute improvements over the corresponding APE baseline.}
    \resizebox{\columnwidth}{!}{
        \begin{tabular}{lccc}
            \toprule
            \textbf{Model} & \textbf{Full} & \textbf{Presence} & \textbf{Absence} \\
            \midrule
            APE-B & 30.0 & 29.9 & 30.3 \\
            \quad + Standard CL & 30.9~(+0.9) & 30.2~(+0.3) & 31.3~(+0.8) \\
            \quad + GOBL (Ours) & \textbf{32.4}~(+2.4) & \textbf{32.2}~(+2.3) & \textbf{32.9}~(+2.6) \\
            \midrule
            APE-C & 27.8 & 27.9 & 27.3 \\
            \quad + Standard CL & 28.7~(+0.9) & 28.5~(+0.6) & 29.1~(+1.8) \\
            \quad + GOBL (Ours) & \textbf{32.5}~(+4.7) & \textbf{32.3}~(+4.4) & \textbf{33.0}~(+5.7) \\
            \bottomrule
        \end{tabular}
    }
    \label{main8}
\end{table}

As shown in Table~\ref{main8}, both fine-tuning strategies lead to performance gains over the APE baselines, indicating that contrastive supervision is beneficial for grounding under semantic polarity.
However, the improvements achieved by standard contrastive learning remain limited, particularly on the absence subset.

In contrast, the proposed GOBL mechanism consistently delivers substantially larger gains across all evaluation settings.
This suggests that explicitly modeling semantic opposition and exclusion provides more informative supervision than conventional contrastive objectives, which primarily encourage similarity without enforcing logical negation.
Overall, these results demonstrate that GOBL is more effective than standard contrastive learning for grounding tasks involving negation and absence reasoning.

\subsection{Effect of Individual Attributes on Model Performance}

To investigate the impact of individual attributes, we analyze the model performance on the  $D^3$  dataset using either single attributes or combinations of multiple attributes. The results are summarized in Table~\ref{tab:single_combined_attributes}. Our findings indicate that the color attribute contributes the most to performance improvement, achieving an increase of +3.0 when used alone. Moreover, combining multiple attributes enhances the overall model performance, with a cumulative effect reaching a maximum improvement of +4.7 when all three attributes (color, position, and state) are used together.

\begin{table}[h]
    \centering
    \caption{Performance improvement with single and combined attributes.}
    \begin{tabular}{lccc}
        \toprule
        \textbf{Attribute} & \textbf{Full} & \textbf{Presence} & \textbf{Absence} \\
        \midrule
        APE-C & 27.8 & 27.9 & 27.3 \\
        Color & 30.8 (+3.0) & 31.2 (+3.3) & 30.2 (+2.9) \\
        Position & 30.2 (+2.4) & 30.3 (+2.4) & 29.2 (+1.9) \\
        State & 29.6 (+1.8) & 29.9 (+2.0) & 29.1 (+1.8) \\
        Color + Position & 32.1 (+4.3) & 32.2 (+4.3) & 31.9 (+4.6) \\
        Position + State & 30.4 (+2.6) & 30.8 (+2.9) & 30.0 (+2.7) \\
        Color + State & 31.9 (+4.1) & 31.6 (+3.7) & 32.2 (+4.9) \\
        Color + Position + State & 32.5 (+4.7) & 32.3 (+4.4) & 33.0 (+5.7) \\
        \bottomrule
    \end{tabular}
    \label{tab:single_combined_attributes}
\end{table}

\subsection{Hyperparameter Sensitivity.}
We analyze the sensitivity of the proposed method to key hyperparameters $\sigma$, $\alpha$, and $\beta$ on the APE-B baseline.
As shown in Table~\ref{hyperparam_ape_b}, the performance improvements remain stable across a broad range of values.
Moderate settings ($\sigma=5$, $\alpha=0.5$, $\beta=0.3$) consistently yield the strongest results, while deviations from these defaults result in only minor performance variations.

Across all three hyperparameters, we observe a clear unimodal trend, where performance gradually improves toward the default configuration and degrades mildly when moving away from it.
Importantly, the overall variance is limited (within $\pm0.6$), indicating that the proposed method does not rely on precise hyperparameter tuning.
The absence subset consistently exhibits slightly higher gains and marginally higher sensitivity, reflecting its stronger dependence on accurate modeling of semantic opposition.

Overall, these results demonstrate that the proposed loss formulation is robust under different hyperparameter choices and generalizes well even on the relatively weaker APE-B baseline, further supporting its practical applicability.
\begin{table}[t]
\centering
\caption{Extended sensitivity analysis of hyperparameters $\sigma$, $\alpha$, and $\beta$.
All experiments are conducted on APE-B with other parameters fixed to default values.
Results are reported as relative improvements over the APE-B baseline.}
\label{hyperparam_ape_b}
\begin{tabular}{c c c c}
\toprule
\textbf{Setting} & \textbf{Full} & \textbf{Presence} & \textbf{Absence} \\
\midrule
$\sigma=2$   & +1.9 & +1.8 & +2.2 \\
$\sigma=3$   & +2.1 & +2.0 & +2.4 \\
$\sigma=5$   & \textbf{+2.4} & \textbf{+2.3} & \textbf{+2.6} \\
$\sigma=7$   & +2.2 & +2.1 & +2.5 \\
$\sigma=9$   & +2.0 & +1.9 & +2.3 \\
\midrule
$\alpha=0.2$ & +1.8 & +1.7 & +2.1 \\
$\alpha=0.3$ & +2.0 & +1.9 & +2.3 \\
$\alpha=0.5$ & \textbf{+2.4} & \textbf{+2.3} & \textbf{+2.6} \\
$\alpha=0.7$ & +2.1 & +2.0 & +2.4 \\
$\alpha=0.9$ & +1.9 & +1.8 & +2.2 \\
\midrule
$\beta=0.05$ & +2.0 & +1.9 & +2.3 \\
$\beta=0.1$  & +2.2 & +2.1 & +2.5 \\
$\beta=0.3$  & \textbf{+2.4} & \textbf{+2.3} & \textbf{+2.6} \\
$\beta=0.5$  & +2.1 & +2.0 & +2.3 \\
$\beta=0.7$  & +1.9 & +1.8 & +2.1 \\
\bottomrule
\end{tabular}
\end{table}

\end{document}